\documentclass[conference]{IEEEtran}
\IEEEoverridecommandlockouts
\usepackage{cite}
\usepackage{amsmath,amssymb,amsfonts}
\usepackage{graphicx}
\usepackage{textcomp}
\usepackage{xcolor}

\usepackage{makecell}
\usepackage{pgfplots}
\def\BibTeX{{\rm B\kern-.05em{\sc i\kern-.025em b}\kern-.08em
    T\kern-.1667em\lower.7ex\hbox{E}\kern-.125emX}}
\begin{document}

\title{End-to-End High Accuracy License Plate Recognition Based on Depthwise Separable Convolution Networks}

\author{
 \IEEEauthorblockN{	Song-Ren Wang, Hong-Yang Shih, Zheng-Yi Shen, and Wen-Kai Tai}
\IEEEauthorblockA{Computer Science and Information Engineering\\ 
	National Taiwan University of Science and Technology\\ 
	Taipei, Taiwan \\
    Email: \{B10415005, B10415045, M10715098, wktai\}@mail.ntust.edu.tw}
}

\maketitle

	\begin{abstract}
	Automatic license plate recognition plays a crucial role in modern
	transportation systems such as for traffic monitoring and vehicle violation
	detection. In real-world scenarios, license plate recognition still faces
	many challenges and is impaired by unpredictable interference such as
	weather or lighting conditions. Many machine learning based ALPR solutions
	have been proposed to solve such challenges in recent years. However, most
	are not convincing, either because their results are evaluated on
	small or simple datasets that lack diverse surroundings, or because they require
	powerful hardware to achieve a reasonable frames-per-second in real-world applications. In
	this paper, we propose a novel segmentation-free framework for license plate
	recognition and introduce NP-ALPR, a diverse and challenging dataset
	which resembles real-world scenarios. The proposed network model consists
	of the latest deep learning methods and state-of-the-art ideas, and benefits
	from a novel network architecture. It achieves higher accuracy with lower
	computational requirements than previous works. We evaluate the effectiveness
	of the proposed method on three different datasets and show a recognition
	accuracy of over 99\% and over 70 fps, demonstrating that our
	method is not only robust but also computationally efficient. 
\end{abstract}

\section{Introduction}

Automatic license plate recognition (ALPR) is an essential part of research
for intelligent transportation systems such
as surveillance systems for access control, traffic monitoring, traffic
violation detection, and parking lot management. 

Despite the large body of approaches proposed for ALPR~\cite{GSHsu,Khalid,AOLP,TKCheang,Rayson,Cheng-Hung,Xie-Fei}, 
challenges still exist in real-world
applications. For example, highly distorted or blurred vehicle images, poor
lighting conditions, and frigid weather conditions all significantly
influence the recognition process. As most previous solutions rely
on extra rules (e.g., the maximum number of characters allowed in a plate) to
enhance their accuracy, or validate their methods on datasets that are
environment-specific (images are collected using a single camera or from identical
viewing angles), or lack diversity (for instance, they recognize only a single
class of vehicles or plates with the same background color), they
perform well only in restricted scenarios. A robust ALPR system, however,
should address these common challenges and adapt to diverse
environmental conditions.

\begin{figure*}[t]
	\centering
	\begin{tabular*}{\linewidth}{@{\extracolsep{\fill}} cc cc cc cc @{}}
		\includegraphics[height=0.13\textwidth,width=0.238\textwidth]{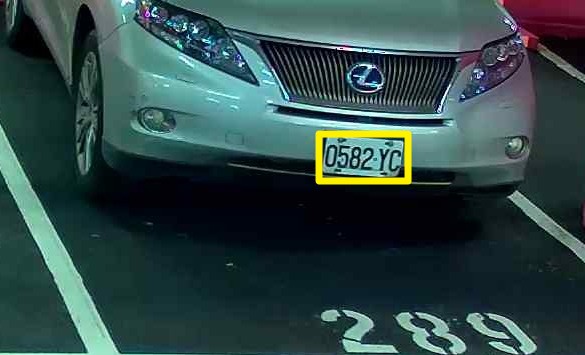} & 
		\includegraphics[height=0.13\textwidth,width=0.238\textwidth]{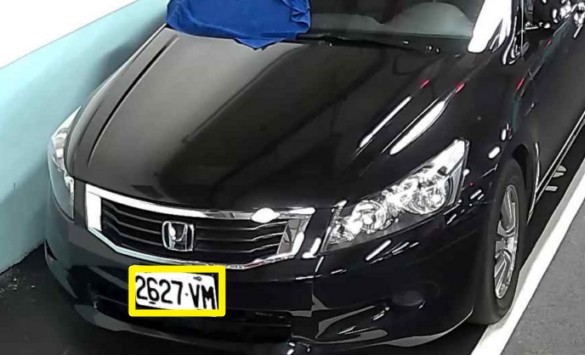} &   \includegraphics[height=0.13\textwidth,width=0.238\textwidth]{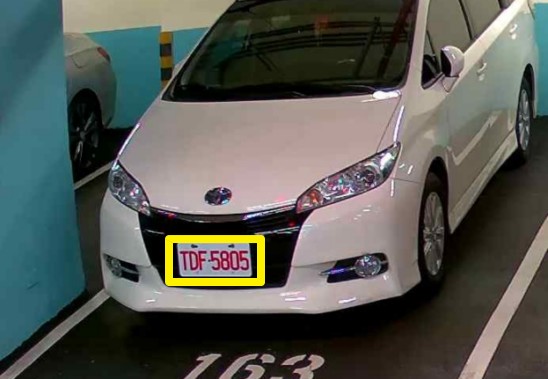} &
		\includegraphics[height=0.13\textwidth,width=0.238\textwidth]{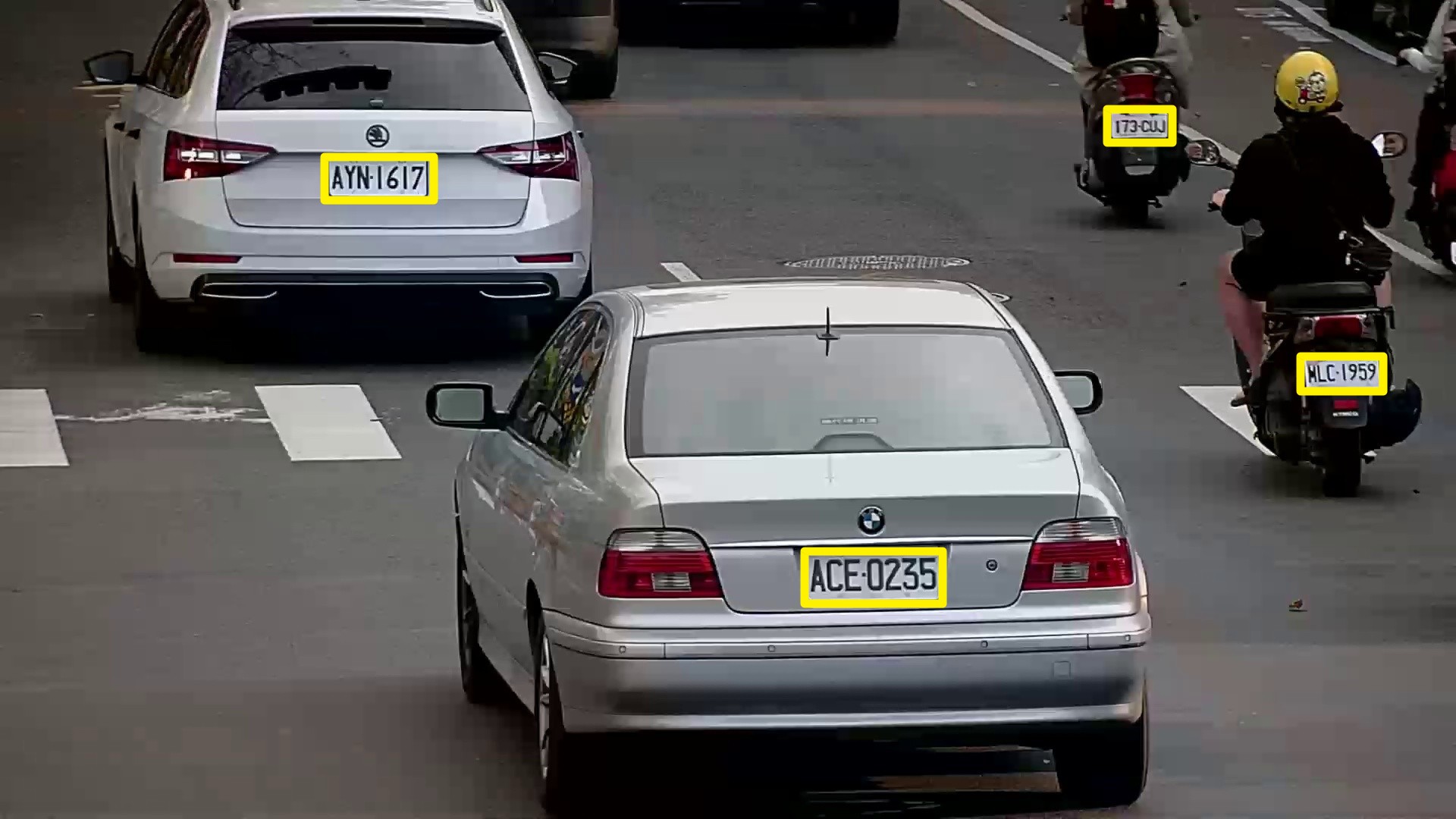} &
		\\
		\includegraphics[height=0.13\textwidth,width=0.238\textwidth]{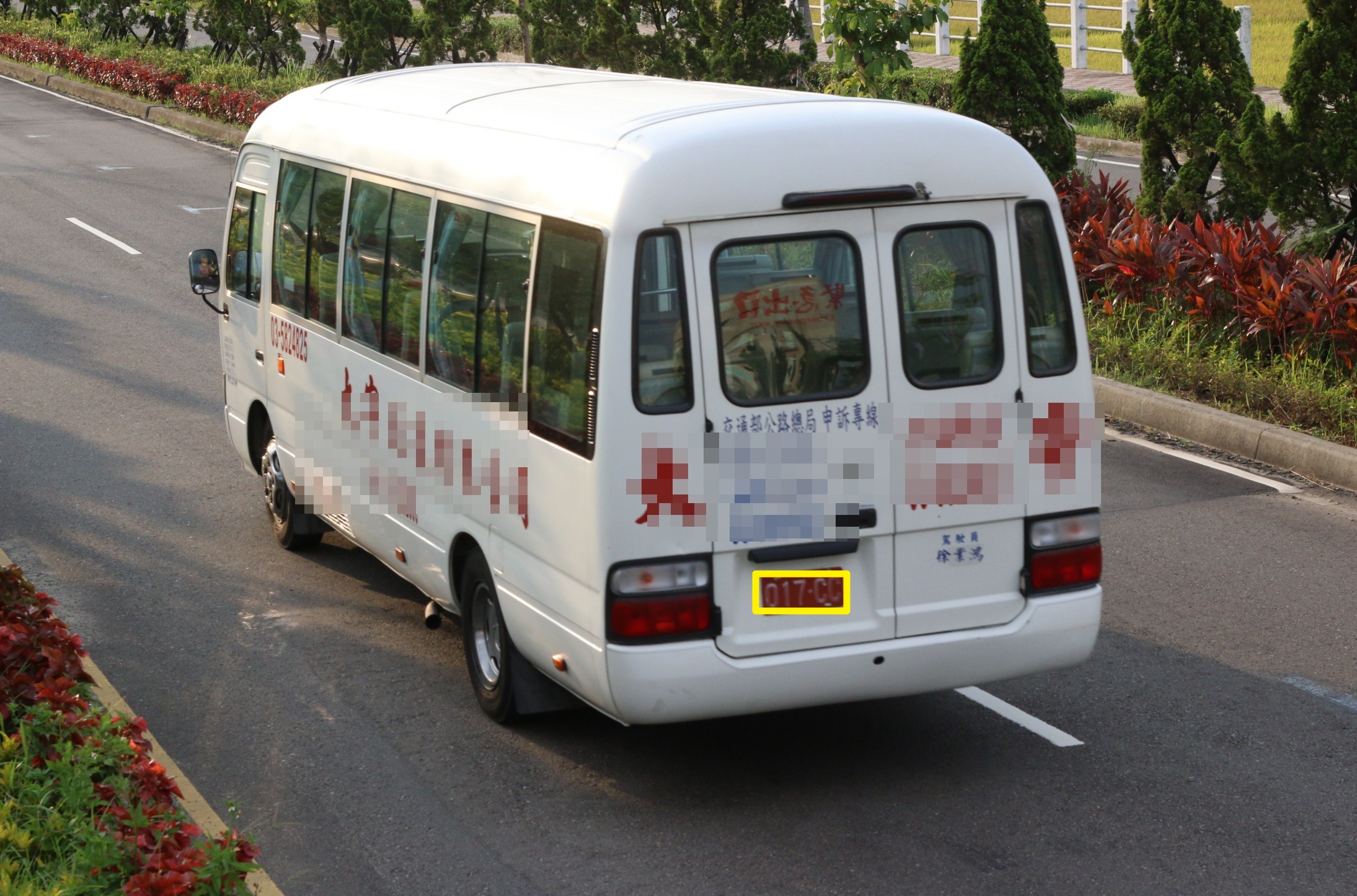} &   \includegraphics[height=0.13\textwidth,width=0.238\textwidth]{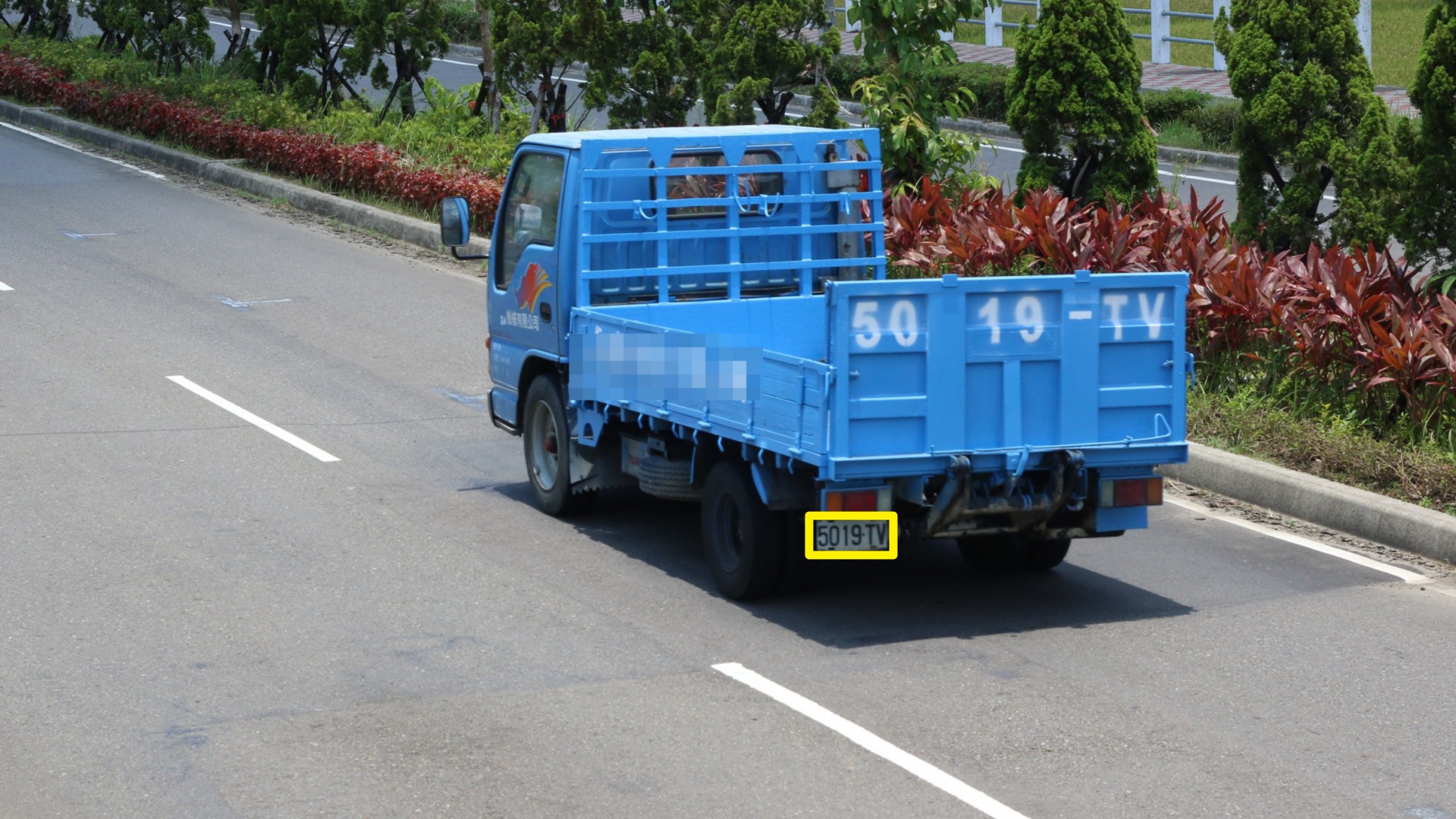} &
		\includegraphics[height=0.13\textwidth,width=0.238\textwidth]{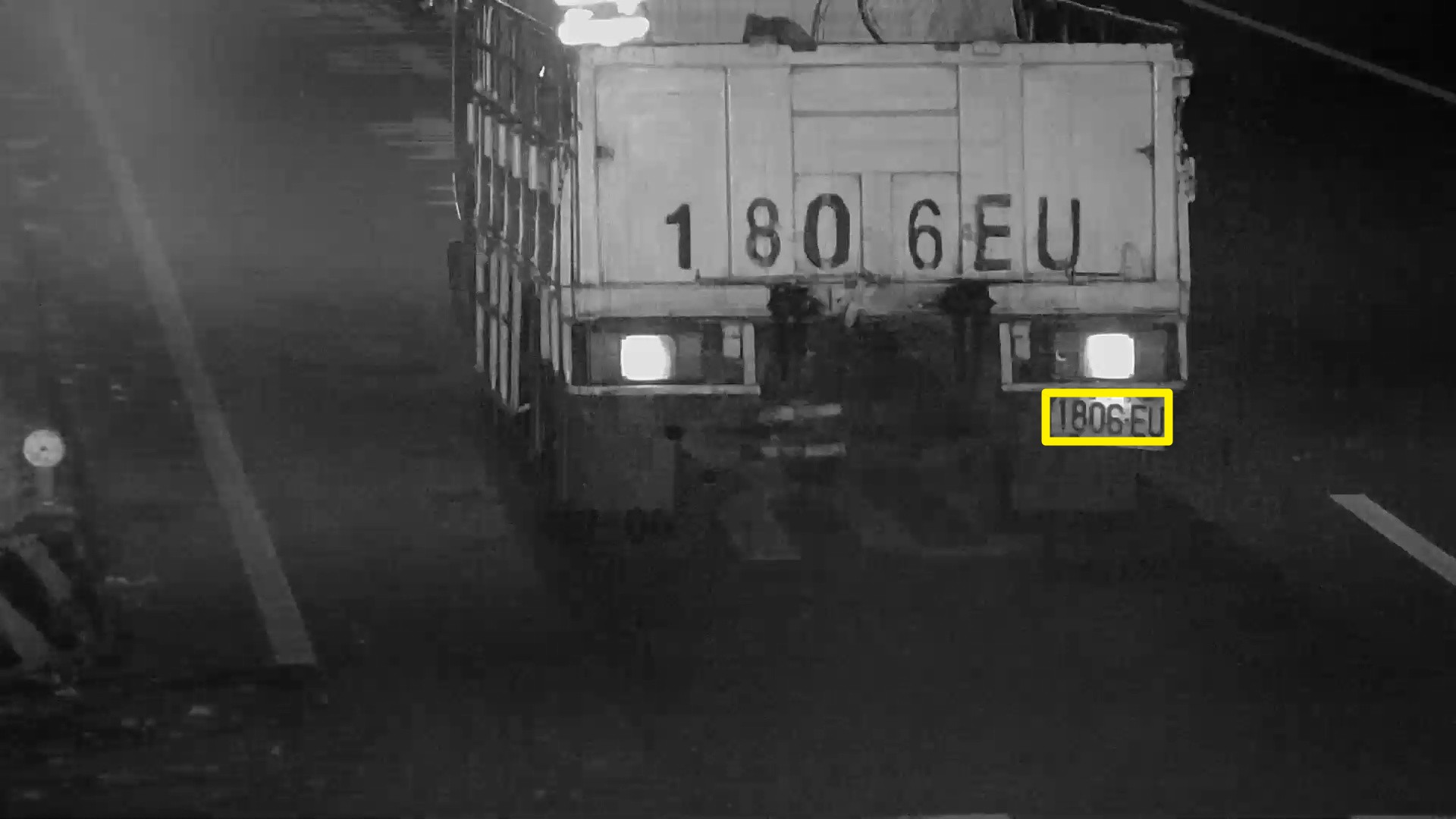} &  \includegraphics[height=0.13\textwidth,width=0.238\textwidth]{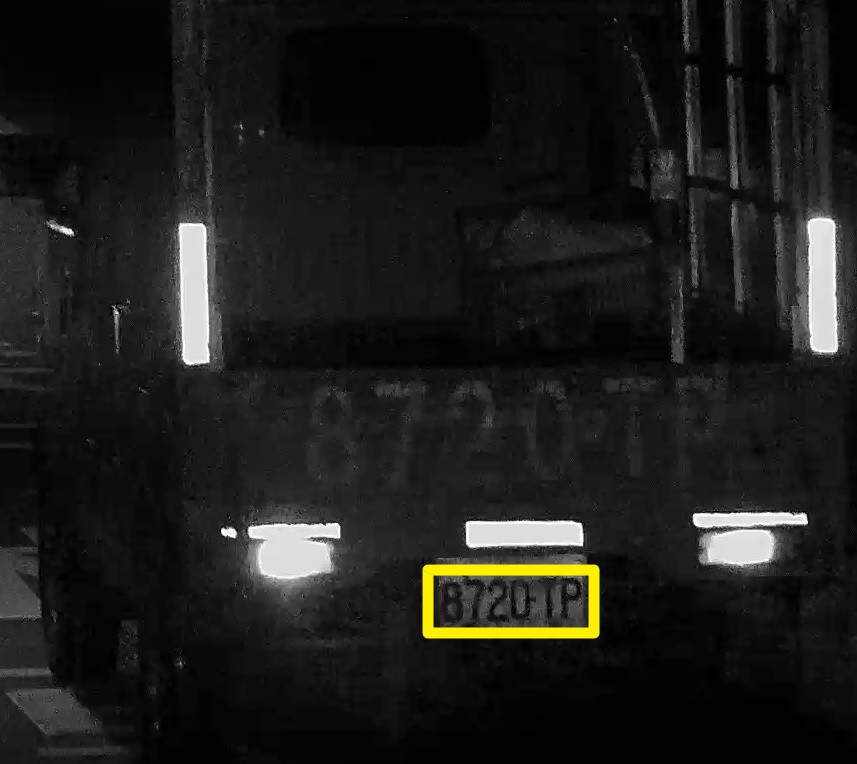} &
		\\
		\includegraphics[height=0.13\textwidth,width=0.238\textwidth]{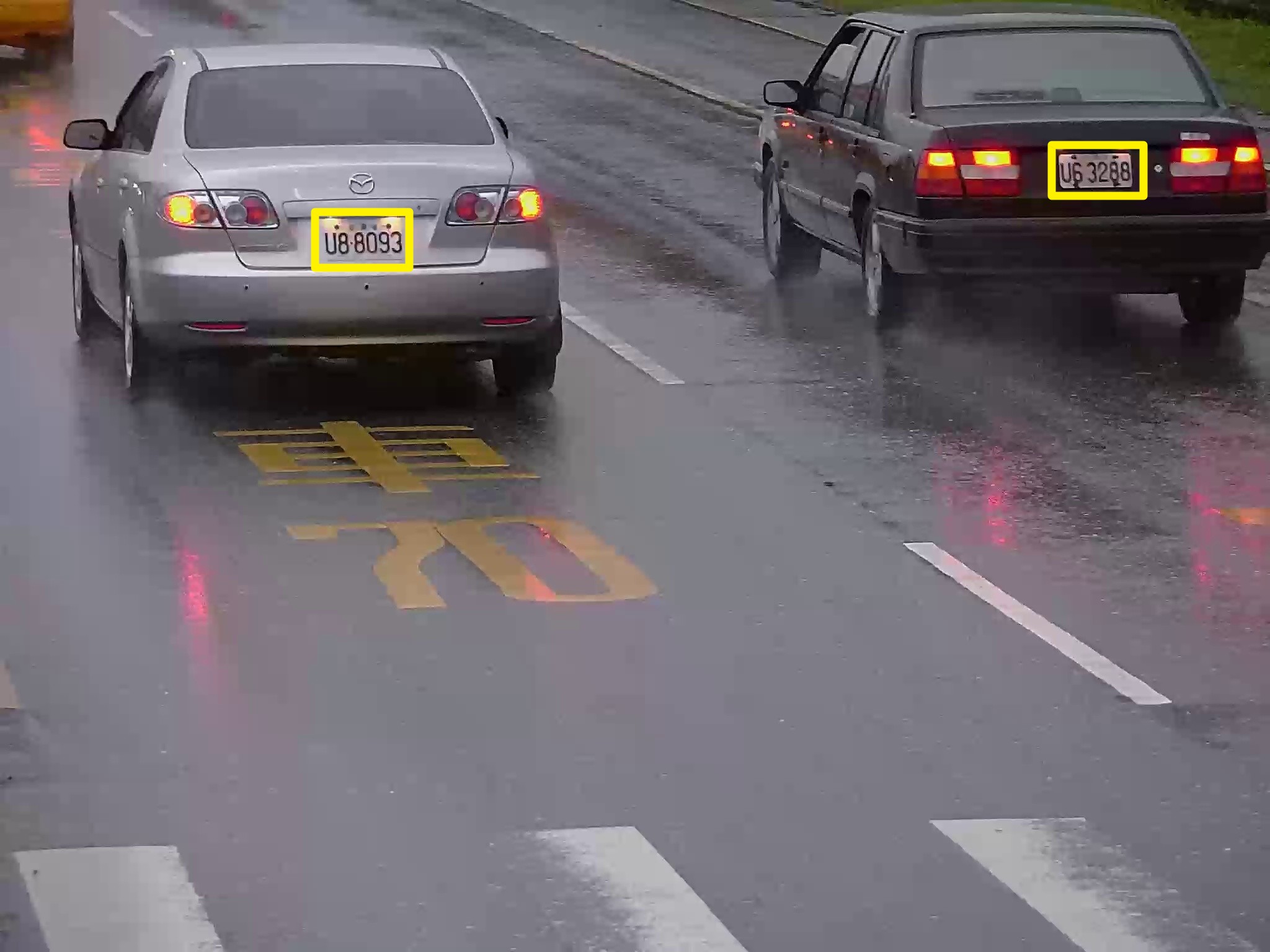} &   \includegraphics[height=0.13\textwidth,width=0.238\textwidth]{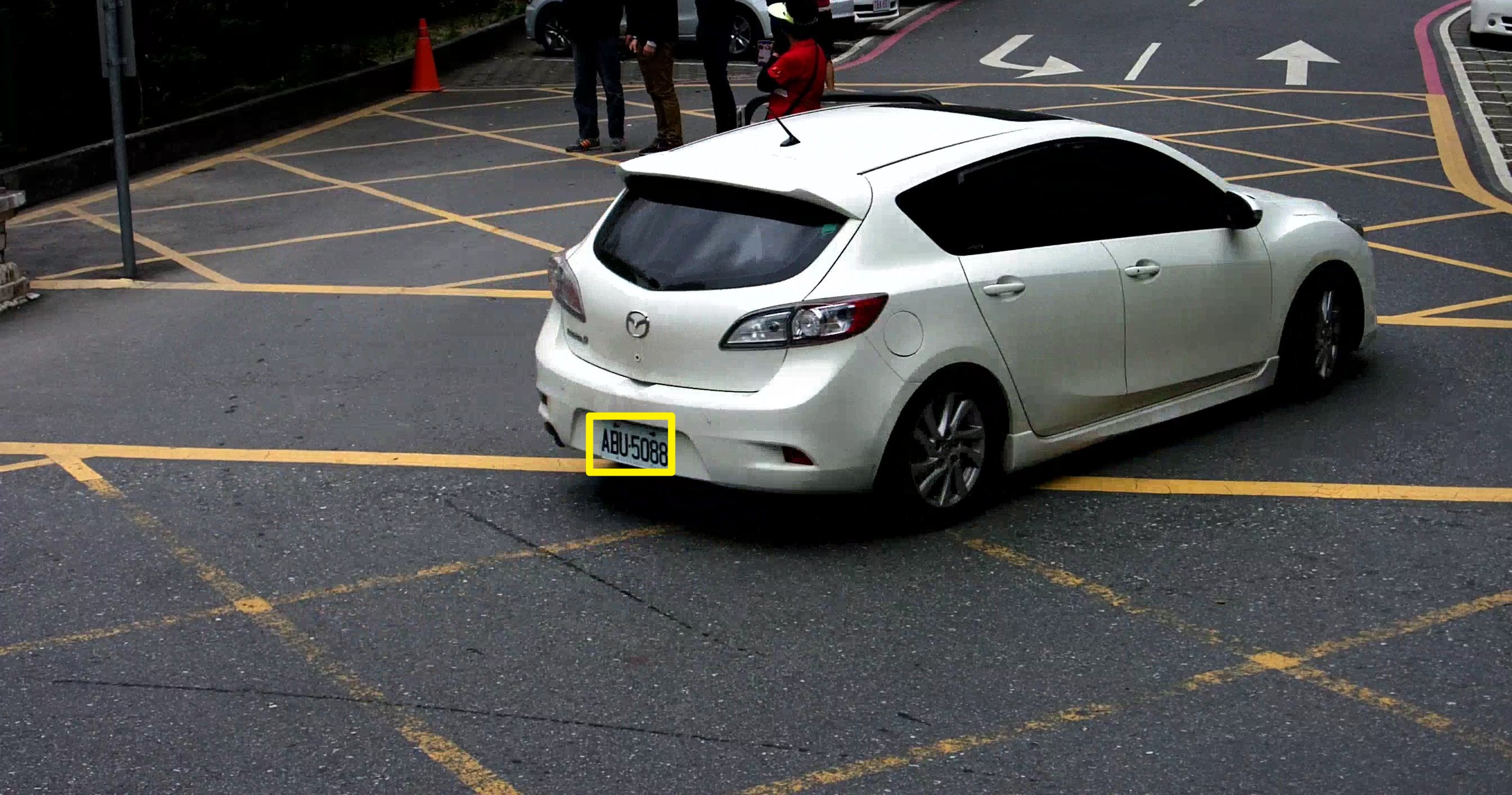} &
		\includegraphics[height=0.13\textwidth,width=0.238\textwidth]{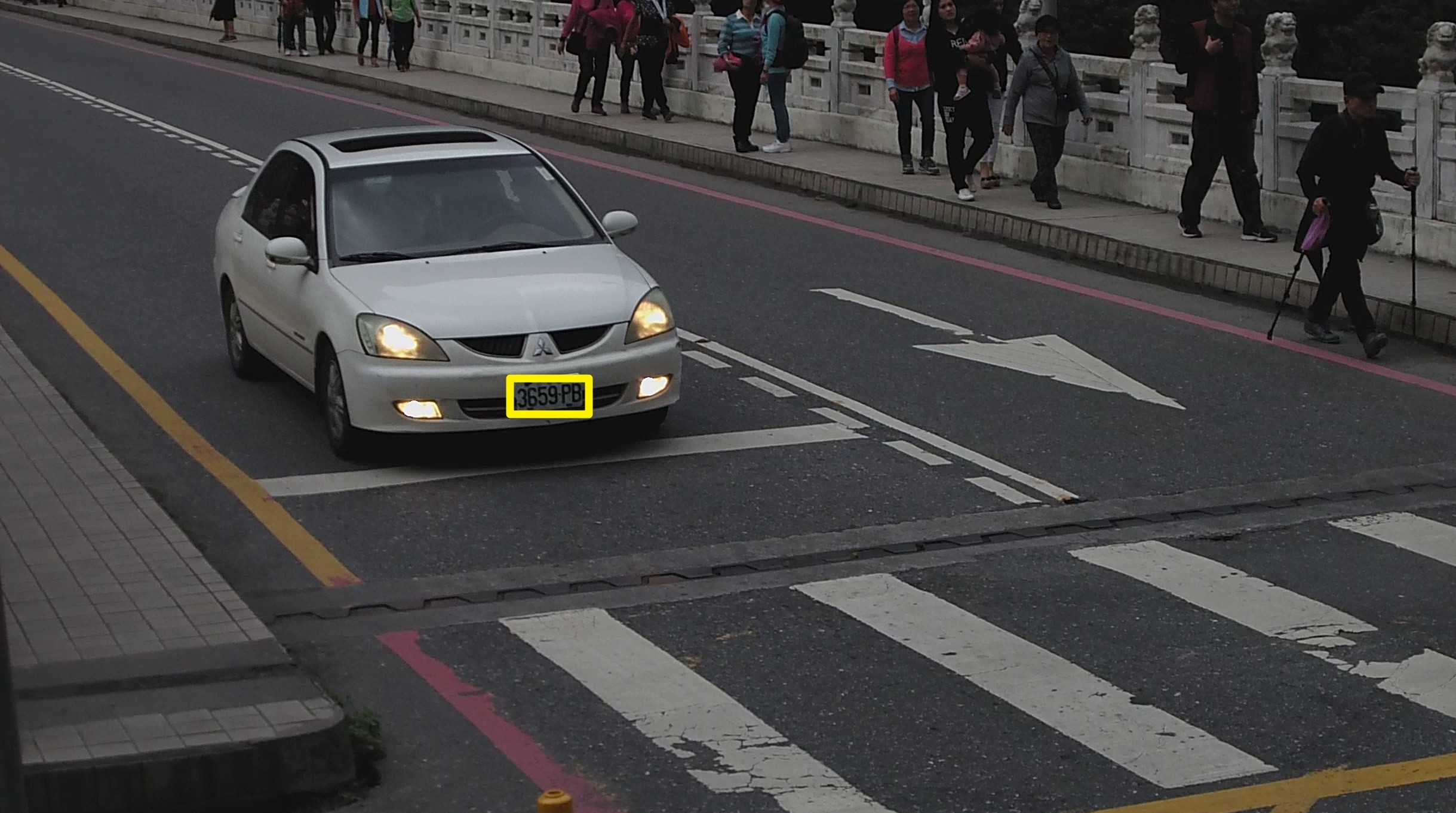} &  \includegraphics[height=0.13\textwidth,width=0.238\textwidth]{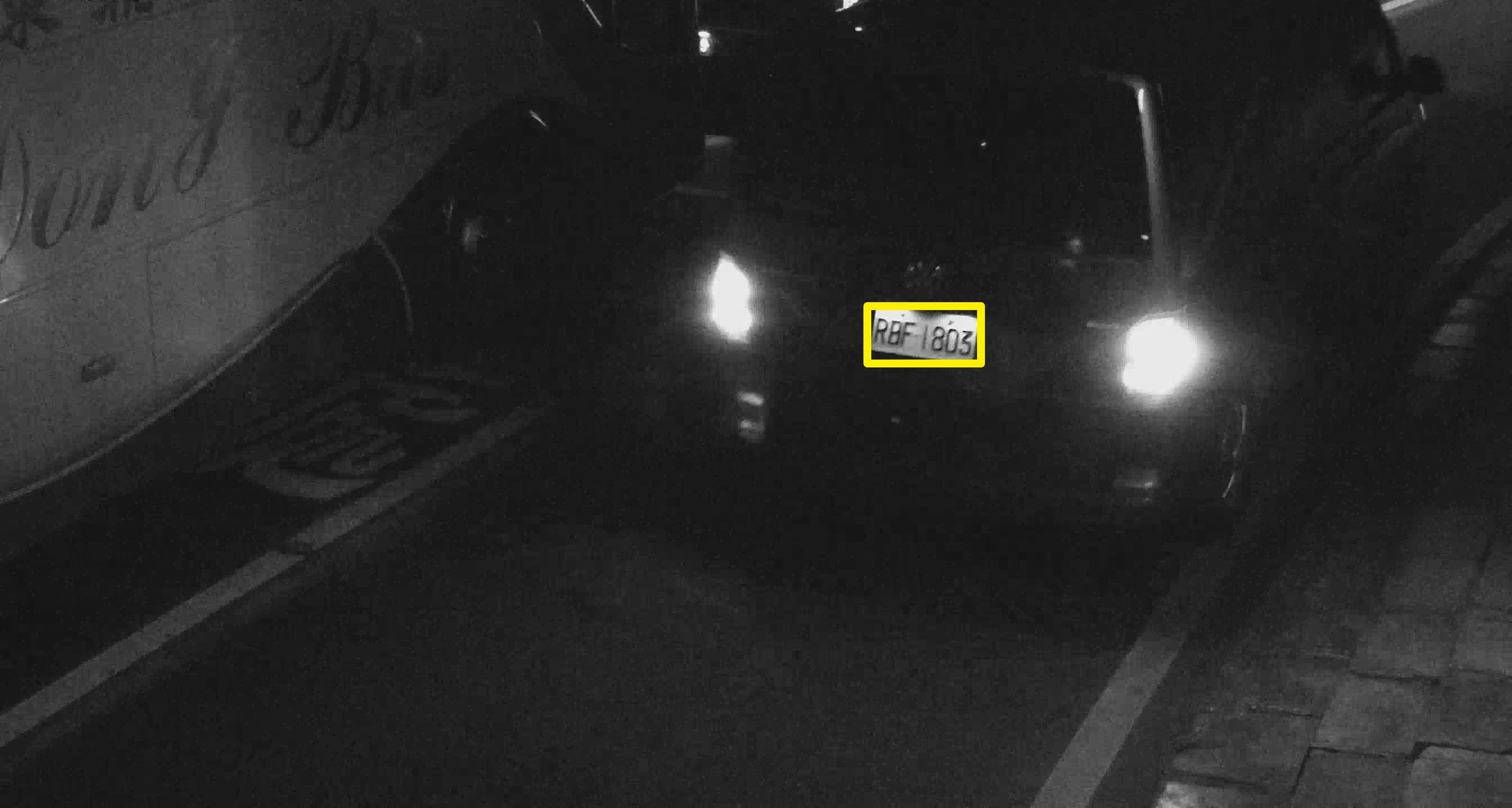} &
		\\
	\end{tabular*}
	\caption{Samples of NP-ALPR dataset showing the dataset variety,
		including different times of day, locations, lighting conditions,
		and various vehicle types and positions. Yellow rectangles represent 
		manually annotated LP locations. Due
		to privacy concerns, sensitive information is blurred.}
	\label{fig:np_dataset}
\end{figure*}

Traditional license plate recognition (LPR) methods are commonly based on
segmentation-based methods, that is, character detection followed by
character recognition. Such methods require accurate character segmentation:
faults in segmentation lead to misrecognition of license plates even when 
using a robust character recognizer. However, in real-world scenarios,
as blurry images and environmental factors degrade the accuracy
of character segmentation, these methods are not suitable for 
real-world applications; moreover, separating character detection and recognition
brings with it additional computational costs. With the
development of deep learning techniques, more and more innovative ideas for
solving LPR without character segmentation have been 
proposed~\cite{TKCheang,Holistic,JZhuang}. Segmentation-free methods
usually extract features from license plates and deliver them to a CNN or RNN
model to recognize the character sequence. Practically, these methods have
better performance and accuracy and are more robust, compared to
segmentation-based methods.

In this paper, we propose a novel segmentation-free neural network
architecture for LPR based on Xception~\cite{Xception} and
Inception-ResNet v2~\cite{Inception-v4}. With the deep learning
infrastructure provided by both, the network learning is deeper
and requires fewer parameters; that is, it is efficient and fast, compared to
other machine learning methods. In addition, we integrate an affine
transformation model into the proposed recognition system to rectify distorted
images before they are submitted to the recognition model. Several datasets
are used to verify the proposed method, showing that it
achieves 99\% accuracy with an average fps (frames per second) of over 75, indicating that it
is applicable under different environments and challenging conditions
without any implementation change.

Apart from using public datasets~\cite{GSHsu,Holistic,Rayson} to validate the proposed method, we also describe 
NP-ALPR,\footnote{Due to local privacy
	policies, we cannot make NP-ALPR publicly available. For access to
	NP-ALPR, please contact the first author.} 
another large dataset with over 10,000
vehicle images recorded from several cameras under various conditions,
including locations, vehicle distance, lighting conditions, and times of day.
Furthermore, this dataset contains various types of vehicles
(motorcycles, cars, buses, and trucks), license plates with various
background colors (white, green, yellow, and red), and license plates with
characters in various colors (black, white, and red). Examples of the
dataset are shown in Figure~\ref{fig:np_dataset}. Note that the dataset
also includes images with multiple vehicles in the same frame.

To summarize, the main contributions made by this work are as follows:

\begin{itemize}
	\item We propose a novel end-to-end LPR model without character segmentation based
	on Xception and Inception ResNet v2.
	
	\item The proposed LPR method recognizes license plate characters
	at 99\% accuracy and over 75 fps under different datasets, confirming
	that it is robust enough to be applied in real-world applications and
	that it outperforms several recent works.
	
	\item We present NP-ALPR, a dataset that contains over 10,000 images of
	various types of vehicles under various conditions.
\end{itemize}

In the next sections, we first discuss approaches raised in recent
years to recognize license plates and then present the
network architecture and training flow of the proposed method. We then
present the experimental results to verify the proposed method.

\section{Related Work}
Common methods for LPR can be divided into 
segmentation-based and segmentation-free methods. In this section, we briefly
review recent work on both LPR methods and
present a brief description of work that our research touches upon.
Since this work mainly focuses on LPR, we do not include studies on
vehicle and license plate detection.

\subsection{Segmentation-based LPR}
Most conventional LPR methods perform character detection first and then
recognize the segemented character (character recognition). Several recent
works~\cite{Rayson,Cheng-Hung,Xie-Fei} are based on this methodology and differ only
in implementation details. 

Existing algorithms for character detection can be classified into two
groups, the first based on connected component techniques that build
connected areas from binarized images and regard each as a 
character~\cite{Khalid,Z-Abderaouf,HLi}, and the other based on
projection techniques that separate each character according to the
top/bottom boundaries obtained from a horizontal projection of binarized
images~\cite{Khalil,Z-Abderaouf,Xie-Fei}. 

Various algorithms have been proposed for character recognition which are
based either on template matching or on machine learning methods. The former
has been used widely for character recognition by measuring the similarity
between the segmented character and template images~\cite{ERLee,MAKo,Khalil}. For the latter, ~\cite{MZahedi,YWang} proposed works that apply
a scale-invariant feature transform (SIFT)~\cite{Abdel-Hakim} to
extract features from segmented characters for delivery to
support vector machines (SVMs)~\cite{Srivastava} for final classification.
As this method requires numerous sliding windows to extract license plate features,
however, it is more computationally costly. The regional
convolutional neural network (RCNN) avoids such unecessary computations by
using selective search to optimize the detection, but it requires more
effort to train the models. Other common machine learning methods include
probabilistic neural networks (PNNs), hidden Markov models (HMMs), and multilayer
perceptrons (MLPs).

\subsection{Segmentation-free LPR method}

Solutions for LPR without character detection emerged following 
advances in machine learning and deep learning. Elimination of
character detection enables these approaches to achieve better performance
relative to segmentation-based methods. \cite{HLi} propose the
first segmentation-free LPR method by using convolutional neural networks
(CNNs). LP image features are extracted via CNNs, and a
recurrent neural network (RNN) with connectionist temporal classification
(CTC) is used to label the sequential data and to classify the character
sequence. Similar CNN-based work includes \cite{DMenotti,TKCheang,Holistic}.

To the best of our knowledge, \cite{JZhuang} proposed the first work
to use advanced semantic segmentation for license plate recognition
based on a modified version of DeepLabv2 ResNet-101. This method is based on
pixel-wise classification, which usually yields more robust accuracy but is
more computationally complex. It achieved outstanding results on the AOLP 
dataset~\cite{GSHsu}.

\subsection{Datasets for ALPR}

The SSIG SegPlate Database~\cite{SSIG}, which contains 2k Brazilian license
plate images, is a commonly used benchmark for ALPR research, but as the
images were collected only on sunny days and from a static camera with
monotonous backgrounds, the data is neither representive nor convincing. The
UFPR-ALPR dataset~\cite{Rayson} contains 4.5k images of different types of
vehicles captured by three different cameras, yet does not include variations such as
daytime differences and weather conditions. The AOLP
dataset~\cite{GSHsu} contains three subsets: Access Control (with 681
images), Traffic Law Enforcement (with 528 images), and Road Patrol (with
611 images). Although each subset contains images under different conditions, such
as various times of day, tilted plates, and different viewing points, the
total number of images is relatively small. In \cite{Holistic}, the author
proposes the ReId dataset, which contains 76k license plates. Although it
contains a large number of images of license plates, it lacks
variations in tilt angels. To the best of our knowledge, ~\cite{ZXu}
propose CCPD, the largest-yet dataset, with over 250k
unique car images in China with detailed annotations under diverse
environmental conditions.

\begin{figure}[!htb]
	\centering
	\begin{tabular}{@{}cccccc@{}}
		\includegraphics[height=0.06\textwidth,width=0.15\textwidth]{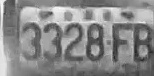} & 
		\includegraphics[height=0.06\textwidth,width=0.15\textwidth]{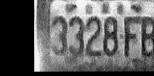} &   \includegraphics[height=0.06\textwidth,width=0.15\textwidth]{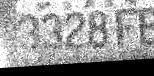} &
		\\
		\includegraphics[height=0.06\textwidth,width=0.15\textwidth]{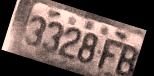} & 
		\includegraphics[height=0.06\textwidth,width=0.15\textwidth]{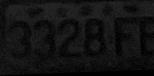} &   \includegraphics[height=0.06\textwidth,width=0.15\textwidth]{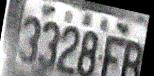} &
		\\
		\includegraphics[height=0.06\textwidth,width=0.15\textwidth]{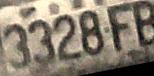} & 
		\includegraphics[height=0.06\textwidth,width=0.15\textwidth]{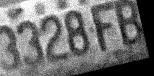} &   \includegraphics[height=0.06\textwidth,width=0.15\textwidth]{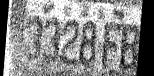} &
		\\
	\end{tabular}
	\caption{Dataset augmentation generated by imgaug library
		with different settings. The left-top image is the raw image.}
	\label{fig:augmentation1}
\end{figure}

\begin{table}[!htb]
	\caption{The detailed architecture descriptions of the proposed network.}\smallskip
	\centering
	\resizebox{.95\columnwidth}{!}{
		\smallskip\begin{tabular}{llll}
			\Xhline{2\arrayrulewidth}\noalign{\smallskip}
			& \makecell{Layer} & \makecell{Input/Parameter} & Output \\
			\noalign{\smallskip}\hline\noalign{\smallskip}
			& Concat & (128, 32),  (128, 32) & (128, 32, 2) \\  
			& Conv + BN + LeakyReLU & (128, 32, 2) & (64, 16, 32) \\ 
			& Conv + BN + LeakyReLU & (64, 16, 32) & (64, 16, 64) \\ 
			& Xception Module & (64, 16, 64) & (64, 16, 64) \\
			& Xception Module & (64, 16, 64) & (64, 16, 64) \\
			& Inception Module B & (64, 16, 64) & (64, 16, 64) \\
			& Inception Module B & (64, 16, 64) & (64, 16, 64) \\
			& Xception Reduce Module & (64, 16, 64) & (32, 8, 128) \\
			& Xception Module & (32, 8, 128) & (32, 8, 128) \\
			& Xception Module & (32, 8, 128) & (32, 8, 128) \\
			& Inception Module B & (32, 8, 128) & (32, 8, 128) \\
			& Inception Module B & (32, 8, 128) & (32, 8, 128) \\
			& Xception Module & (32, 8, 128) & (32, 8, 128) \\
			& Xception Module & (32, 8, 128) & (32, 8, 128) \\
			& Permute & (32, 8, 128) & (8, 32, 128) \\
			& GlobalAvgPool1D & (8, 32, 128) & (32, 128) \\
			& Dropout & 0.4 ratio &  \\
			& LSTM & (32, 128) & (32, 38) \\
			& BatchNorm &  &  \\
			& Softmax &  &  \\
			\noalign{\smallskip}\Xhline{2\arrayrulewidth}
	\end{tabular}}
	\label{tab:network_setting}       
\end{table}

\begin{table} [!htb]
	\caption{All possible combinations of Taiwan license plate; where 'A' denotes an arbitrary alphabet, 'N' denotes an arbitrary number character.}\smallskip
	\centering
	\resizebox{.95\columnwidth}{!}{
		\smallskip\begin{tabular}{lcccccccc}
			\Xhline{2\arrayrulewidth}\noalign{\smallskip}
			& \makecell{LP with \\ 4 characters} & \makecell{LP with \\ 5 characters} & \makecell{LP with \\ 6 characters} & \makecell{LP with \\ 7 characters} \\
			\noalign{\smallskip}\hline\noalign{\smallskip}
			& AA-AN & AA-ANN & AA-NNNN & AAA-NNNN \\
			& AA-NN & AA-NNN & AN-NNNN &  \\
			& AN-NN & AN-NNN & NA-NNNN &  \\
			& NA-NN & NA-NNN & AAA-NNN &  \\
			& NN-AA & NNN-AA & AAN-NNN & \\
			& NN-AN & NNN-AN & ANA-NNN & \\
			& NN-NA & NNN-NA & ANN-NNN &  \\
			&  &             & NAA-NNN &  \\
			&  &             & NNN-AAA &  \\
			&  &             & NNNN-AA &  \\
			&  &             & NNNN-AN &  \\
			\noalign{\smallskip}\Xhline{2\arrayrulewidth}
		\end{tabular}
	}
	\label{tab:lp_rule}       
\end{table}

\section{Proposed Method}

In this section, we present the architectural overview and data
flow of the entire network, and also describe the training process and
provide implementation details. 

\subsection{Overview}
The main purpose of this work is to introduce a real-time and highly
accurate license plate recognition method. Unlike segmentation-based methods
which require character segmentation followed by character
recognition to predict the license character sequence, the proposed model
processes the whole license plate image without segmentation. We also
use CTC loss for segmentation-free training so that we do not need to
annotate the positions of characters in license plates. Here RGB license
plate images cropped from raw images are considered as the inputs for the
model rather than the raw images with vehicles and other background
material. Practically, this is easily accomplished with various detection
solutions. For example, previous work~\cite{Rayson} uses 
YOLOv2~\cite{YoloV2} or the latest YOLOv3~\cite{YoloV3} as the license plate detector and
vehicle detector. In this work, we simply use YOLOv3 as a license plate
detector in several experiments that require license plate
detection. Since this work mainly focuses on license plate recognition,
details on detection methods are not presented here. 

\begin{figure*}[!htb]
	\centering
	\begin{tabular*}{\linewidth}{@{\extracolsep{\fill}} cc cc cc cc @{}}
		\includegraphics[height=0.50\textwidth,width=1.02\textwidth]{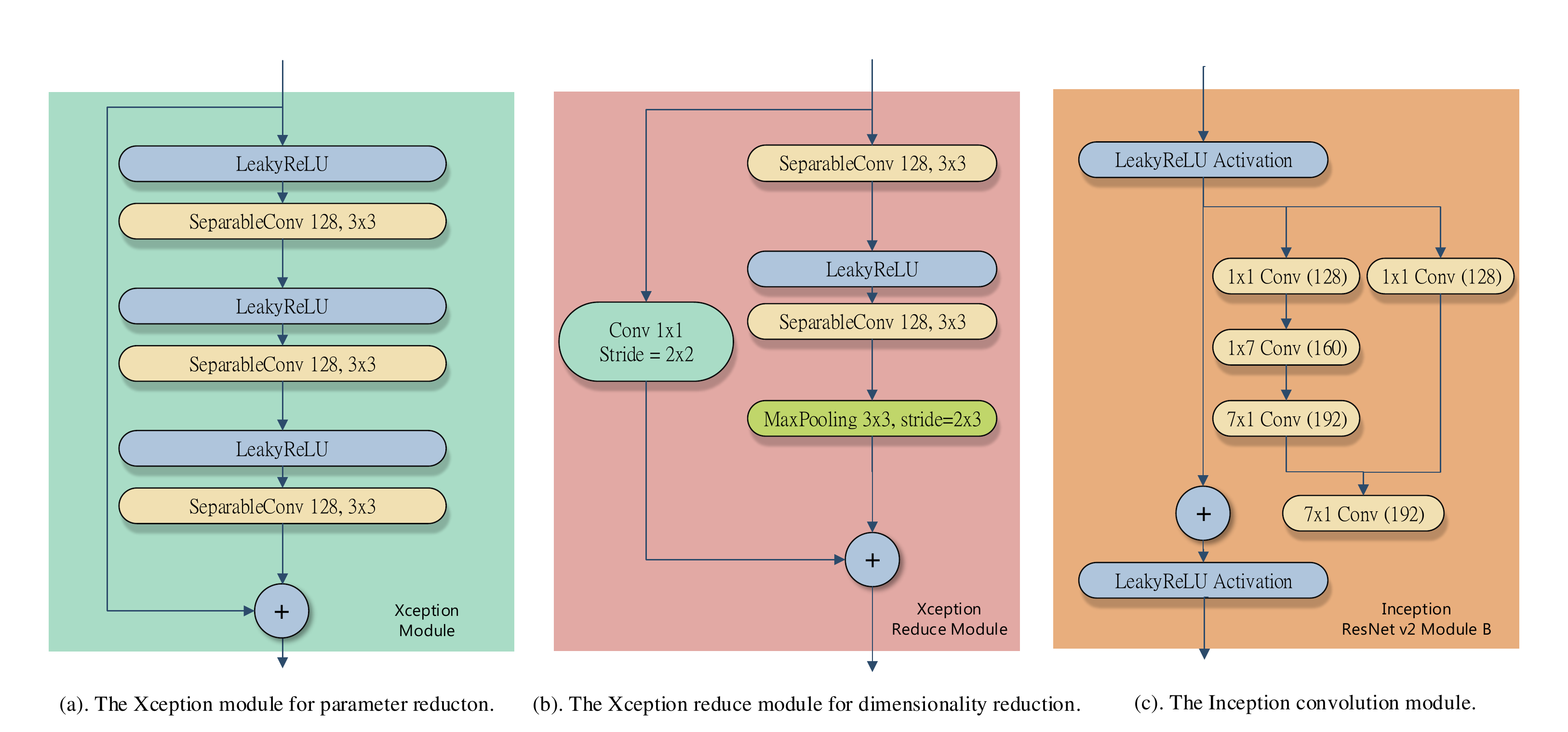} 
	\end{tabular*}
	\caption{Implementation details of Xception and Inception building
		blocks originally proposed in \cite{Xception,Inception-v4}. As names of these building blocks were not given, 
		we name them according to their functionalities. Also, we adopt 
		LeakyReLU as the activation function,
		whereas in the original work they did not reveal the implementation details.}
	\label{fig:googLENet}
\end{figure*}

Existing powerful networks such as AlexNet, VGGNet, or GoogLeNet are widely
popular in recent work. However, to
build a fast and lightweight network, 
wholesale use of    these monolithic networks    
is not the best option. In this work, the basic building blocks of
the proposed networks were inspired either by Inception ResNet 
v2~\cite{Inception-v4} or Xception~\cite{Xception}, or both. The Inception
network is a deep neural network that achieves outstanding performance
with a modest number of parameters; due to its complicated design, however, 
it is still too computationally costly for use in license plate recognition 
applications. In \cite{Inception-v4}, improved versions of Inception with
residual connections prevent vanishing gradients; that is,
Inception ResNet v1 and v2 achieve slightly better
performance than their predecessors. On the other hand, the Xception
network, although not as powerful as the former, benefits
from the adoption of depthwise separable convolution and is thus significantly
more efficient and requires fewer parameters to match the
performance of the former. To strike a good balance between
accuracy and computational efficiency, we make use of portions of both to
construct a license plate recognition-specific network rather than directly
use both heavy networks. The building blocks used to construct
license plate recognition model are illustrated in Figure~\ref{fig:googLENet}.
Note that in the original works~\cite{Xception,Inception-v4}, few
implementation details are given, and ReLU is used for
activation in both works. However, in several experiments, we found that
replacing ReLU with LeakyReLU improved overall efficiency. Hence
in this work we use LeakyReLU as the activation function.

\begin{figure*}[!htb]
	\centering
	\begin{tabular*}{\linewidth}{@{\extracolsep{\fill}} cc cc cc cc @{}}
		\includegraphics[height=0.38\textwidth,width=0.5\textwidth]{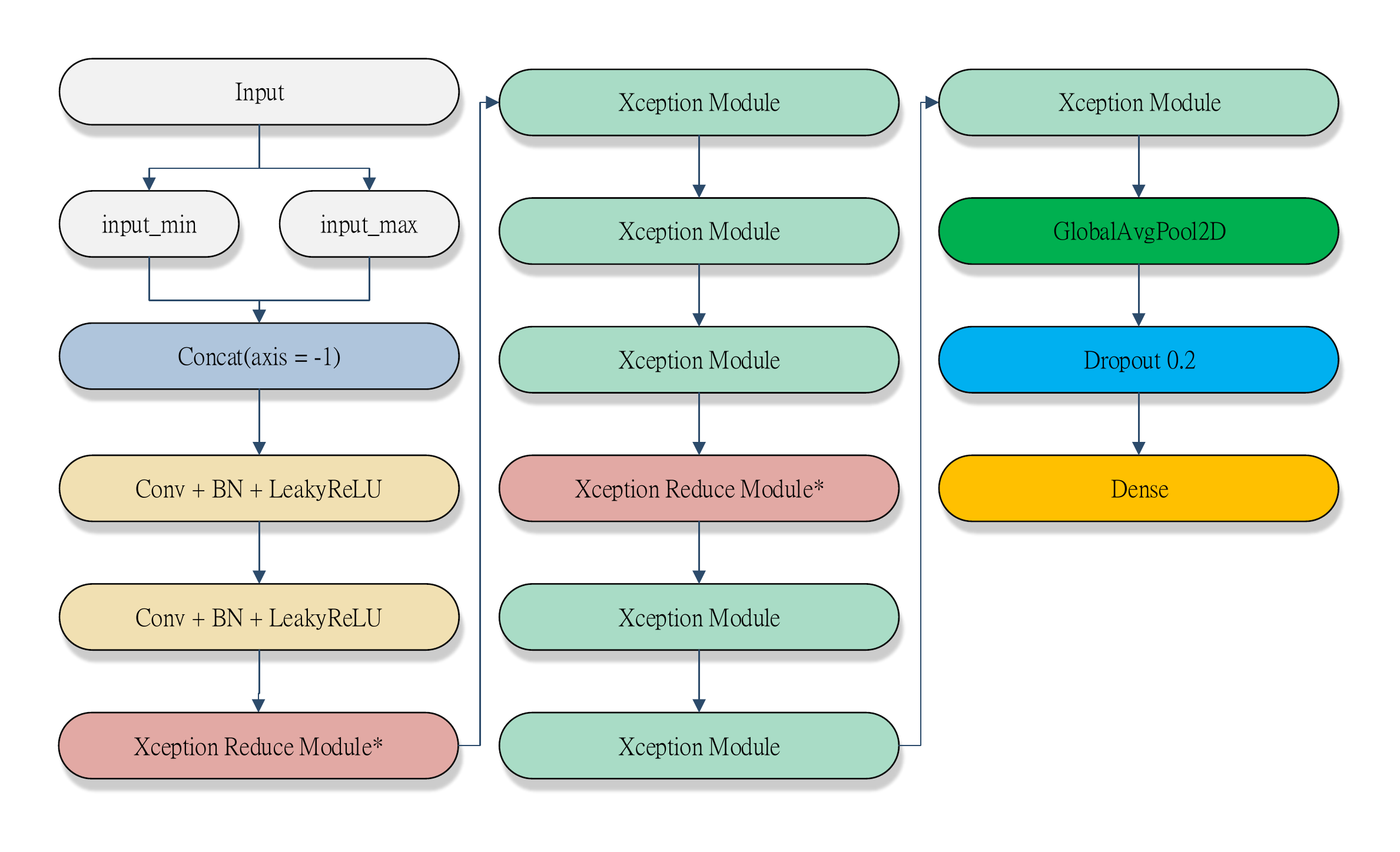} &
		\includegraphics[height=0.58\textwidth,width=0.525\textwidth]{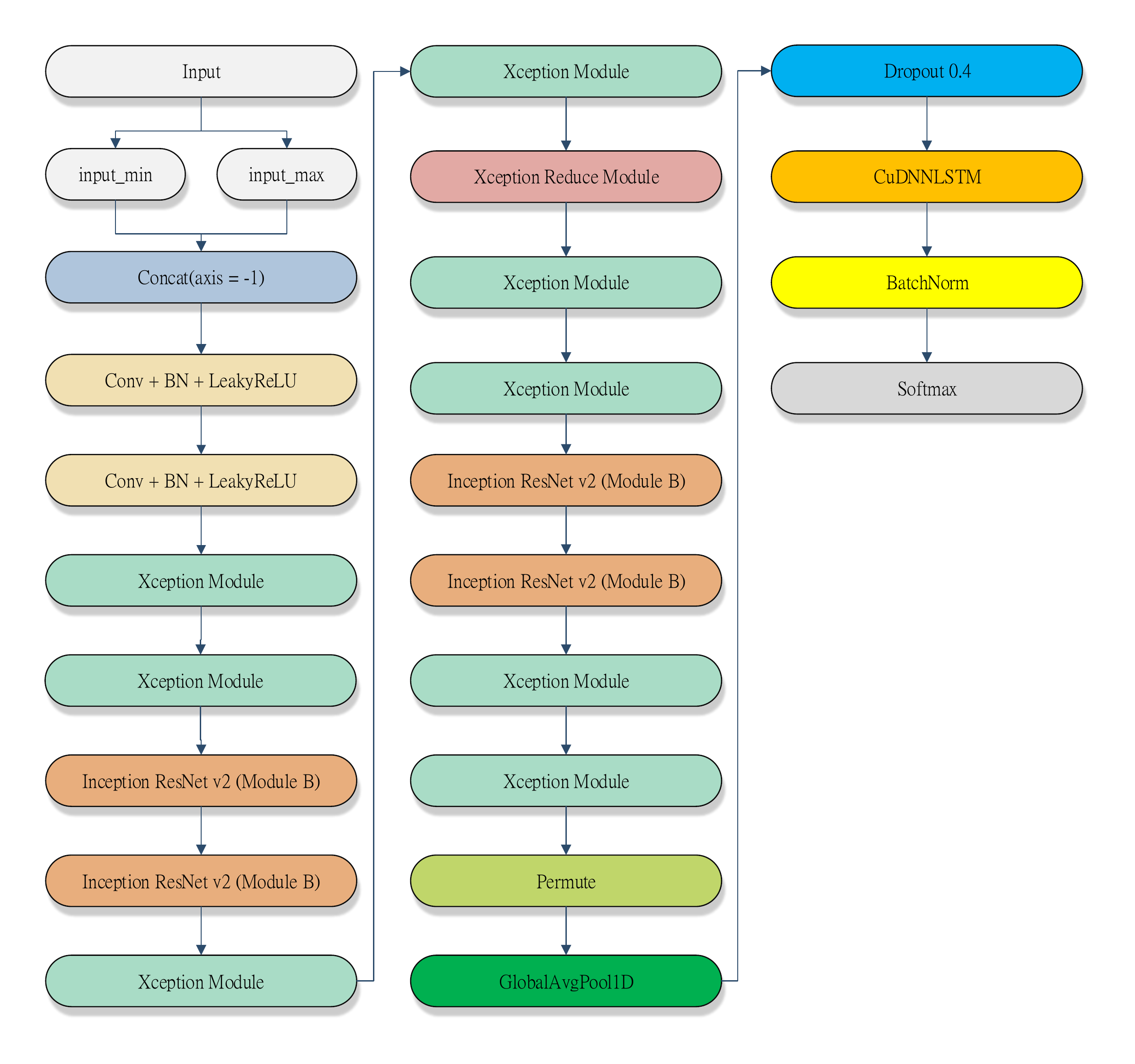} 
		\\ (a). Affine transformation model & (b). Proposed license
		plate recognition model
	\end{tabular*}
	\caption{Detailed overview of proposed neural network
		architecture, including affine transformation model and license plate
		recognition model.}
	\label{fig:network_overview}
\end{figure*}

\subsection{Affine Transformation Module}
Tilted license plate images are common in real-world scenarios and 
often lead to inaccurate prediction. To take this into account, we correct the tilt
of input images before the license plate recognition phase.
To rectify tilted license plate images, we apply affine transformation to
unwarp them, so that in the license plate recognition phase, every license plate
resembles one captured from the frontal view. To precisely unwarp
distorted license plate images, we train an affine transformation model to
capture the license plate's four endpoints and deliver them to the correction
algorithm. We experimented with several affine transformation algorithms,
and found the one provided by the OpenCV library to be 
more stable than others. Thus we adopted it in our implementation. 
Figure~\ref{fig:network_overview} shows the detailed architecture of the
affine transformation model. The results show that the 
affine transform model yields a roughly 1 to 3 percent increase in recognition
accuracy.

\subsection{License Plate Recognition Module}

The network architecture of the proposed segmentation-free license plate
recognition model is presented in Figure~\ref{fig:network_overview}. To begin
with, image features are extracted using a pre-trained CNN model that
slides across a license plate bounding box with an input size of 128 $\times$ 32,
and principal component analysis (PCA) is applied to reduce the feature
dimensions, yielding feature maps of size 32 $\times$ 8, which are then 
delivered to the DNN model with a bidirectional
CudnnLSTM layer followed by a fully connected layer to obtain sequential
features. Lastly, CTC is applied to decode the sequential LSTM features
and to predict the character sequence. The parameters of the
network architecture are shown in Table~\ref{tab:network_setting}.

During the experiments, we observed that although higher input sizes yielded more
accurate results, we found that $128 \times 32$ is the best choice for
the network as a trade-off between accuracy and performance.
To increase efficiency, we also attempted to reduce the feature map size to 
$16 \times 4$, but this led to significantly inaccurate results.

\subsection{Model Training}
Accurate prediction usually requires model training on a large number of samples.
However, existing datasets are commonly small-scale and thus do not
satisfy this requirement. To take this into account, we first trained
our model using an autoencoder, an unsupervised learning algorithm that allows
the model be trained on only a few labeled samples, after which transfer
learning is using to enhance the model. Secondly, we made use of the \texttt{imgaug} library to
augment the training samples by generating blurred and distorted images. Examples
produced by the \texttt{imgaug} library are shown in Figure~\ref{fig:augmentation1}.
Further, because some license plate letters are relatively infrequent
in our region (for example, I and O), we synthesized infrequent characters
to produce fabricated license plate images and used them to train the model.
To do so, we first randomly chose segregated characters 
cropped from the dataset, and then measured the similarity between
the currently selected character and the partly completed license plate to determine
whether the character should be kept or discarded. To make the
fabricated license plate resemble a real one, we followed all legal
rules of character combinations of license plates shown in 
Table~\ref{tab:lp_rule} to fabricate license plates. The experiment results showed
that fabricated samples have a character distribution similar to real
license plates, and that they benefit not only
accuracy for license plates with infrequent characters but also overall performance.

\section{Experiment}
In this section, we present our experimental results 
on the AOLP dataset, the UFPR-ALPR dataset, and our own NP
dataset. Although the CCPD dataset includes a huge number of license plate
images, because non-ASCII characters are used to represent the province
code, we did not evaluate on this dataset as our model currently does not 
support non-ASCII characters. Here we consider both recognition accuracy and
running time as performance metrics and provide several comparisons with
state-of-the-art methods.

All of our models were trained using an NVIDIA GeForce RTX 2080 Ti with specific
images provided by each dataset; testing was conducted on a GTX 1070.

\subsection{AOLP Dataset}

\begin{table}
	\caption{AOLP results}	
	\centering
	\resizebox{.95\columnwidth}{!}{
		\smallskip\begin{tabular}{lcccccccc}
			\Xhline{2\arrayrulewidth}\noalign{\smallskip}
			& \makecell{AC} (\%)  & \makecell{LE} (\%)  & \makecell{RP} (\%) & Avg (\%) & fps\\ \noalign{\smallskip}\hline\noalign{\smallskip}
			Hsu (2013) \cite{AOLP} & 88.50 & 86.60 & 85.70 & 86.93 & 7.00 & \\
			LSTMs (2016) \cite{HLi} & 94.85 & 94.19 & 88.38 & 92.47 & None & \\
			DeepFCN (2016) \cite{FCN} & 97.90 & 97.60 & 98.20 & 97.90 & 69.40 & \\
			Zhuang (2018) \cite{JZhuang} & 99.41 & 99.31 & 99.02 & 99.25 & 38.00 & \\
			Proposed & 99.13 & 99.20 & 99.21 & 99.18 & 78.41 & \\
			\noalign{\smallskip}\Xhline{2\arrayrulewidth}
	\end{tabular}}
	\label{tab:aolp_result}       
\end{table}

\begin{table}
	\centering
	\caption{AOLP error study}	
	\centering
	\resizebox{.95\columnwidth}{!}{
		\smallskip\begin{tabular}{lcccccccc}
			\Xhline{2\arrayrulewidth}\noalign{\smallskip}
			\makecell{Image}  & \makecell{Ground truth}  & \makecell{Predicted} & Cause of failure\\ \noalign{\smallskip}\hline\noalign{\smallskip}
			\makecell{\includegraphics[height=0.05\textwidth,width=0.15\textwidth]{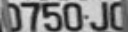}} & 0750J0 & 0750J\textcolor{red}{C} & \makecell{Character `0' \\ covered and \\ incomplete.}  & \\
			\makecell{\includegraphics[height=0.06\textwidth,width=0.15\textwidth]{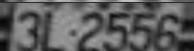}} & 3L2556 & 3L2556\textcolor{red}{5} & \makecell{Background \\ treated as part\\ of license plate.}  & \\
			\makecell{\includegraphics[height=0.06\textwidth,width=0.15\textwidth]{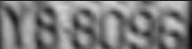}} & Y88096 & Y8\textcolor{red}{B}096 &  \makecell{Blurry image \\ leads to prediction \\ failure.} & \\
			\makecell{\includegraphics[height=0.05\textwidth,width=0.15\textwidth]{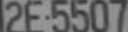}} & 2E5507 & 2\textcolor{red}{F}5507 &  \makecell{`E' character \\ is incomplete.} & \\
			\noalign{\smallskip}\Xhline{2\arrayrulewidth}
	\end{tabular}}
	\label{tab:aolp_error_study}       
\end{table}

The application-oriented license plate benchmark dataset (AOLP) includes
2,049 Taiwan license plate images and is divided into access control (AC), law
enforcement (LE), and road patrol (RP) scenarios. In this dataset, of the three 
subsets, the LE subset seems to be the most challenging and most similar to 
real-world scenarios. The LE subset contains 757 images of vehicles violating
traffic laws that were captured from roadside cameras under various
illumination and weather conditions; the image backgrounds 	are cluttered 
with road signs and sometimes single frames even contain multiple 
license plates. The AC and RP subsets, in contrast, are more or less rigid scenarios 
in that most vehicles are very close to the camera. 

In this experiment, we compared the proposed method with four approaches: 
~\cite{AOLP}, ~\cite{HLi}, ~\cite{FCN}, and 
~\cite{JZhuang}. We followed the same training/test split as in 
~\cite{HLi} and~\cite{JZhuang}: two subsets for training
and one for testing. The experimental results are shown in 
Table~\ref{tab:aolp_result}. They show that not only does the recognition
accuracy of the proposed method outperform three previous works \cite{AOLP},
\cite{HLi} and~\cite{FCN}, but the recognition speed is also faster than these
methods, even on less powerful hardware such as the GTX 1070. While the error rate of
the proposed work is slightly higher than that of ~\cite{JZhuang}, our running
speed is twice as fast, which suggests that there is still room for improvement.
For example, we could sacrifice running time
by using a heavier or more complicated model for more accurate predictions. 
In Table~\ref{tab:aolp_error_study} we analyze a few failure cases.
Recognition failures are caused mainly by poor image quality, as mentioned in \cite{JZhuang}.
Moreover, the dataset contains images that are difficult even for human beings 
to recognize, such as the first sample in the table, in which the first and 
last characters are blocked by the surroundings.

\subsection{UFPR-ALPR Dataset}
The UFPR dataset consists of 4.5k images: 1.8k for testing,
1.8k for training, and 900 for validation. In this dataset, images extracted from videos that are captured
from static cameras are divided into different subsets, each of which
contains 30 images with only one vehicle. Actually, every image is a
single frame of the original video, so the background has no significant
change compared to the AOLP dataset. We conducted experiments with the
same settings as in \cite{Rayson} except that we used YOLOv3 for vehicle and
license plate detection, unlike the previous experiment which requires only
license plate recognition. This dataset is thus not as challenging
as the AOLP dataset, and image quality is also far better than that
of AOLP (images in the UFPR dataset are $1920 \times 1080$,
whereas those in AOLP are around $320 \times 480$ or even
lower), but to verify the robustness and the ability of environment
adoptions, 
we still evaluate the method on various datasets.

\begin{table}
	\centering
	\caption{UFPR results}	
	\resizebox{.95\columnwidth}{!}{
		\smallskip\begin{tabular}{l l cc cc}
			\Xhline{2\arrayrulewidth}\noalign{\smallskip}
			& \makecell{Accuracy}(\%) & & \makecell{fps}  \\ \noalign{\smallskip}\hline\noalign{\smallskip}
			\cite{Rayson} & 78.33 (47/60)& & 35 & \\
			Sightound \cite{Rayson} & 70.00 (72/60)& & None & \\
			OpenALPR \cite{Rayson} & 56.67 (34/60)& & None & \\
			Proposed & 76.66 (46/60) & & 43  & \\
			\noalign{\smallskip}\Xhline{2\arrayrulewidth}
		\end{tabular}
	}
	\label{tab:ufpr_result}       
\end{table}

\begin{table}
	\centering
	\caption{UFPR error study}
	\centering
	\resizebox{.95\columnwidth}{!}{
		\smallskip\begin{tabular}{lcccccc}
			\Xhline{2\arrayrulewidth}\noalign{\smallskip}
			\makecell{Image}  & \makecell{Ground truth}  & \makecell{Predicted} & Cause of failure\\ \noalign{\smallskip}\hline\noalign{\smallskip}
			\makecell{\includegraphics[height=0.05\textwidth,width=0.15\textwidth]{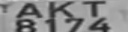}} & \makecell{AKT \\ 8174} & \textcolor{red}{0G27} & \makecell{Completely failed \\ on multiple-row \\ license plates.}  & \\  
			\makecell{\includegraphics[height=0.06\textwidth,width=0.15\textwidth]{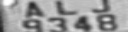}} &  \makecell{ALJ \\ 9348} & \textcolor{red}{B7Q2} & \makecell{Completely failed \\ on multiple-row\\ license plates.}  & \\  
			\makecell{\includegraphics[height=0.05\textwidth,width=0.15\textwidth]{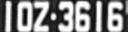}} & IOZ3616 & I\textcolor{red}{0}Z3616 &  \makecell{`O' character \\ extremely \\ similar to `0'.} & \\
			\makecell{\includegraphics[height=0.06\textwidth,width=0.15\textwidth]{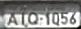}} & AIQ1Q56 & A\textcolor{red}{T}Q1Q56 &  \makecell{`I' character \\ is shadowed.} & \\
			\noalign{\smallskip}\Xhline{2\arrayrulewidth}
	\end{tabular}}
	\label{tab:ufpr_error_study}       
\end{table}

In this experiment, we compared the proposed method with \cite{Rayson}. Because
each subset contains images of the same vehicle, we could simply use
the majority vote to produce the final prediction result of every subset as in
\cite{Rayson}. The results are shown in Table~\ref{tab:ufpr_result}: at first glance,
it appears that our method performs worse than \cite{Rayson}.
After carefully inspection, we find that our prediction model does not
correctly predict license plates with two rows. Of the 60 different
vehicles in the testing set, 12 motorcycles are all equipped with
two-row license plates. In our network design, as the CTC model treats the
license plate as a collection of characters within a single row, the current
proposed method does not fully recognize license plates that have multiple
rows. Again, we demonstrate and analyze failed samples in 
Table~\ref{tab:ufpr_error_study}.

\begin{table}
	\centering
	\caption{Secondary UFPR results}	\centering
	\resizebox{.95\columnwidth}{!}{
		\smallskip\begin{tabular}{lcccccccc}
			\Xhline{2\arrayrulewidth}\noalign{\smallskip}
			& \makecell{Accuracy (\%)} & & \makecell{fps}  \\
			\noalign{\smallskip}\hline\noalign{\smallskip}
			Proposed (without majority vote)& 86.13 (1292/1500)& & 45.24  & \\
			Proposed (with majority vote)& 94.00 (47/50)& & 44.57  & \\
			\makecell[l]{Proposed (without object \\ detection and majority vote)}& 93.13 (1397/1500)& & 69.75  & \\  
			\noalign{\smallskip}\Xhline{2\arrayrulewidth}
	\end{tabular}}
	\label{tab:second_ufpr_result}       
\end{table}

Due to time constraints, we were unable to modify the proposed method to account for
multiple-row license plates. We instead conducted another experiment
on the same dataset but without the motorcycles. We randomly chose 50 subsets
that consisted only of images of non-motorcycle vehicles from the entire dataset, with
the remaining images for training. The results are shown in 
Table~\ref{tab:second_ufpr_result}: the results without
multiple-row license plates indicate that the proposed method is still robust and
sufficiently accurate under various environments in different countries, although
it is currently limited to single-row license plates. Due to the
segmentation-free recognition architecture and the latest YOLOv3 improvements,
the proposed method clearly achieves a significantly higher fps than
\cite{Rayson}. If we entirely preclude license plate detection, the
proposed method achieves 94.00\% accuracy, whereas \cite{Rayson} yields 95.97\%
and 90.37\% accuracies on character segmentation and recognition
respectively. Thus the proposed method indeed outperforms their work.

\begin{table}
	\centering
	\caption{NP-ALPR error study}
	\resizebox{.95\columnwidth}{!}{
		\smallskip\begin{tabular}{lcccccccc}
			\Xhline{2\arrayrulewidth}\noalign{\smallskip}
			\makecell{Image}  & \makecell{Ground truth}  & \makecell{Predicted} & Cause of failure\\ \noalign{\smallskip}\hline\noalign{\smallskip}
			\makecell{\includegraphics[height=0.05\textwidth,width=0.15\textwidth]{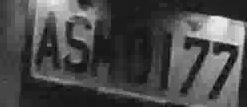}} & \makecell{ASM0177} & ASM\textcolor{red}{B}177 & \makecell{`0' character \\ distorted \\ and shadowed.}  & \\
			\makecell{\includegraphics[height=0.06\textwidth,width=0.15\textwidth]{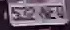}} &  \makecell{532NEQ} & 532\textcolor{red}{A}EQ & \makecell{Image is \\ distorted.}  & \\
			\makecell{\includegraphics[height=0.05\textwidth,width=0.15\textwidth]{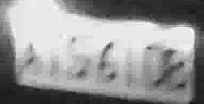}} & A1561D8 & A1561\textcolor{red}{0B} &  \makecell{Image is \\blurred and tilted} & \\
			\makecell{\includegraphics[height=0.06\textwidth,width=0.15\textwidth]{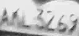}} & AKL3269 & A\textcolor{red}{A}L3269 &  \makecell{Image is \\ extremely blurred.} & \\
			\makecell{\includegraphics[height=0.06\textwidth,width=0.15\textwidth]{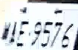}} & MAE9576 & M\textcolor{red}{1}E9576 &  \makecell{Image is \\ extremely blurred.} & \\
			\makecell{\includegraphics[height=0.06\textwidth,width=0.15\textwidth]{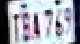}} & T6A769 & T\textcolor{red}{B}A769 &  \makecell{Image is \\ extremely blurred.} & \\
			\makecell{\includegraphics[height=0.06\textwidth,width=0.15\textwidth]{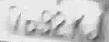}} & 7692KJ & 7\textcolor{red}{0}92KJ &  \makecell{Image is \\ extremely blurred.} & \\
			\makecell{\includegraphics[height=0.06\textwidth,width=0.15\textwidth]{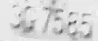}} & 3G7585 & 3\textcolor{red}{6}7585 &  \makecell{Image is \\ extremely blurred.} & \\
			\noalign{\smallskip}\Xhline{2\arrayrulewidth}
	\end{tabular}}
	\label{tab:np_error_study}       
\end{table}

\subsection{NP-ALPR Dataset}
The last experiment was conducted on the proposed NP-ALPR dataset. Because our
model was already pre-trained on a similar private dataset, for this experiment we randomly
selected 1500 images on which to evaluate the proposed method. As
mentioned in the previous section, as diverse conditions such as weather conditions
and times of day are covered in this dataset, this would be the most
practical test in this paper. The proposed method correctly
recognized 1479 license plates (98.60\%) at more than 70 fps on a GTX 1070
GPU. Analysis of the failed samples is given in 
Table~\ref{tab:np_error_study}. In the experiment, we observe that most error
cases were at night, during which rear lights tend to overexpose
cameras, thus increasing the difficulty of license plate recognition. Another difficulty is the
long range and tilted angles between the camera and vehicles; also, the captured
images are often seriously blurred or distorted despite the large $1920 \times 1080$ images.

We note that these extremely distorted and blurred samples are difficult
even for humans to fully recognize. Nevertheless, the proposed
method still achieves over 98.60\% recognition accuracy,
which is outstanding in and of itself.

\section{Conclusion}
In this paper, we present a novel real-time network architecture
combined with the latest works which not only facilitates highly efficient recognition
for deployment in real-world applications but also yields improved overall accuracy. We evaluate
the proposed method on three different datasets and compare the results
with state-of-the-art approaches, showing that it achieves outstanding recognition
accuracy and fast running times. Even though our method does not significantly
outperfom all previous works, it runs on cheaper hardware such as the GTX
1070 GPU, unlike most previous works which use the GTX 1080 TI or higher-level
hardware. As future work, we intend to introduce deblurring
algorithms into our method to improve accuracy under environments such as that in
Table~\ref{tab:np_error_study}, and plan to further optimize the affine
transformation model to reduce the entire running time and errors caused by
incorrect tilting correction. In addition, we plan to recognize multiple-row
license plates to support scenarios such as that in UFPR-ALPR. Finally, we also
intend to support recognition of license plates with non-ASCII characters
such as those in China and Japan.

\bibliographystyle{IEEEtran}
\bibliography{biblio}

\end{document}